%% file: main.tex
\documentclass[10pt,twocolumn,letterpaper]{article}
\input{preamble}
\begin{document}
\maketitle
\input{sec/0_abstract}
\input{sec/0_motivation}
\input{sec/1_intro}
\input{sec/2_related}
\input{sec/3_method}
\input{sec/4_experiments}
\input{sec/5_analysis}
\input{sec/6_conclusion}
\input{sec/7_llm_statement}
{\small
\bibliographystyle{ieeenat_fullname}
\bibliography{references}
}
\end{document}

%% file: preamble.tex
\usepackage[pagenumbers]{cvpr}
\usepackage{amsmath,amssymb,booktabs,array,multirow,colortbl}
\usepackage[T1]{fontenc}
\newcommand{\method}{LoopFit\xspace}
\newcommand{\sg}{\operatorname{sg}}
\definecolor{cvprblue}{rgb}{0.21,0.49,0.74}
\usepackage[breaklinks,colorlinks,allcolors=cvprblue]{hyperref}
\hypersetup{pdftitle={LoopFit: Retrofitting Pretrained Diffusion Transformers with Recurrent Refinement},pdfauthor={Xiaocheng Lu, Haoxi Li, Dazhao Du, Tao Han, Jian Liu, Hao Chen, Bohai Gu, Siyu Yan, Jingcai Guo, Jie Zhang, Song Guo}}
\title{LoopFit: Retrofitting Pretrained Diffusion Transformers\\with Recurrent Refinement}
\author{Xiaocheng Lu\textsuperscript{1}\quad Haoxi Li\textsuperscript{1}\quad Dazhao Du\textsuperscript{1}\quad Tao Han\textsuperscript{1}\quad Jian Liu\textsuperscript{1}\quad Hao Chen\textsuperscript{1}\\[2pt]Bohai Gu\textsuperscript{1}\quad Siyu Yan\textsuperscript{1}\quad Jingcai Guo\textsuperscript{2}\quad Jie Zhang\textsuperscript{1}\quad Song Guo\textsuperscript{1}\\[5pt]
{\normalsize\textsuperscript{1}The Hong Kong University of Science and Technology}\qquad
{\normalsize\textsuperscript{2}The Hong Kong Polytechnic University}}

%% file: sec/0_abstract.tex
\begin{abstract}
Scaling inference computation offers a way to improve diffusion transformers (DiTs), but taking more sampling steps leaves the depth of each denoising prediction unchanged.
Recurrently reusing Transformer blocks increases computation within each step, allowing further refinement of hidden representations.
Retrofitting pretrained models enables this capability without repeating pretraining, but requires blocks learned for fixed-depth computation to refine states produced by their own repeated application.
The challenge is to use each additional pass to correct local errors without disrupting the structure established by earlier passes.
We introduce \textbf{\method}, which uses feedback from earlier passes to guide refinement at each image patch.
Residual Attention Modulation (RAM) uses each patch's feature change in one pass to scale and shift its input to attention in the next.
During training, Residual Relay Loss (RRL) gives greater weight to patches whose prediction errors persist across passes.
A noise-dependent depth schedule speeds up sampling by using four passes at high noise and one at lower noise levels.
At comparable cost, backbones trained with four passes solve more problems using recurrent depth than extra sampling steps.
Across 12 visual puzzle settings, \method solves 89.8\% of boards, compared with 87.8\% for the strongest baseline evaluated on all 12. Most of the gain comes from tracing long paths on mazes and frozen lakes.
\end{abstract}

%% file: sec/0_motivation.tex
\begin{figure*}[t]
  \centering
  \includegraphics[width=0.86\linewidth]{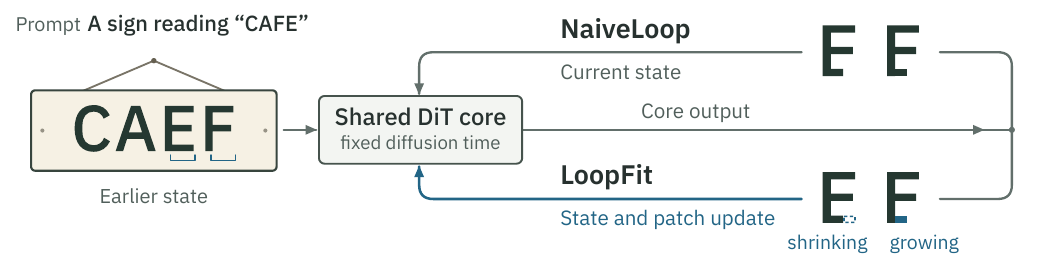}
  \caption{\textbf{What returns to the shared DiT core?}
  The schematic follows the last two letters of the sign. Their bottom strokes reach the same length as one shrinks (dashed outline) and the other grows (solid blue).
  Both models reuse the current state at fixed diffusion time. \method also conditions the next pass on each patch's signed feature change.}
  \label{fig:motivation}
\end{figure*}

%% file: sec/1_intro.tex
\section{Introduction}
\label{sec:intro}

Diffusion transformers (DiTs) generate an image through a sequence of denoising predictions~\cite{ho2020ddpm,peebles2023dit}. Taking more sampling steps gives the model more opportunities to update the image, but each prediction still traverses the same fixed-depth network. Reusing a group of Transformer blocks within a step gives the model more depth to refine its representation of the current noisy image before the next sampling step~\cite{chng2026loopeddit,skiers2026path}. Retrofitting a pretrained DiT with such a loop would let it draw on the image representations it has already learned without repeating pretraining.

The difficulty is that pretraining gives each block a fixed place in the network. At initialization, a single pass through the reused blocks reproduces the pretrained computation. On the next pass, those blocks read their own output. The noisy image and diffusion time stay fixed while neighboring patches can require different corrections. In Fig.~\ref{fig:motivation}, correcting \textsf{CAEF} to \textsf{CAFE} requires shortening the third letter's bottom stroke and extending the fourth's. Midway through, the strokes are equally long. Time and pass index alone give both patches the same conditioning.

Prior work trains recurrent models by supervising intermediate predictions, regulating attention, or specifying how representations should evolve across passes~\cite{chng2026loopeddit,derakhshani2026lift,goyal2026elt,wang2026loopreg}. In a direct DiT retrofit, supervising every pass gives the model a target at each depth, but uniform patch weights give no extra emphasis to errors that survive earlier attempts at correction. In a uniformly supervised loop, we find that the same parts of an image can still carry elevated errors after four passes (Fig.~\ref{fig:t2i-patch-persistence}). A patch's feature update records how the model changed its representation. During training, the target reveals which patches still need correction after that update.

We introduce \textbf{\method}, which carries both forms of feedback across passes. \emph{Residual Attention Modulation} (RAM) retains the signed feature update at each patch and uses it to scale and shift that patch's input to attention on the next pass, building on residual-conditioned modulation~\cite{anonymous2026lrfeed}. Neighboring patches can thus receive different modulation even at the same noise level and recurrent depth. \emph{Residual Relay Loss} (RRL) uses the training target to track each patch's prediction errors across preceding passes and gives more weight to patches with persistently large errors. At inference, only the feature updates are needed. The sampler can allocate these passes across noise levels. At a fixed pass budget, does extra depth help more early or late in denoising?

\method averages 89.8\% success across 12 settings on FrozenLake, Maze, TSP and Sudoku, with the largest gains over the strongest baseline on long paths (Table~\ref{tab:puzzleref}).

Our contributions are threefold.
\begin{itemize}
  \item RAM uses each patch's signed feature update to modulate its attention input on the next pass.
  \item RRL gives greater weight to patches whose prediction errors persist across passes.
  \item A noise-dependent depth schedule reserves extra passes for high-noise steps and uses a single pass at lower noise levels.
\end{itemize}

%% file: sec/2_related.tex
\section{Related Work}
\label{sec:related}
\begin{figure*}[t]
  \centering
  \includegraphics[width=\linewidth]{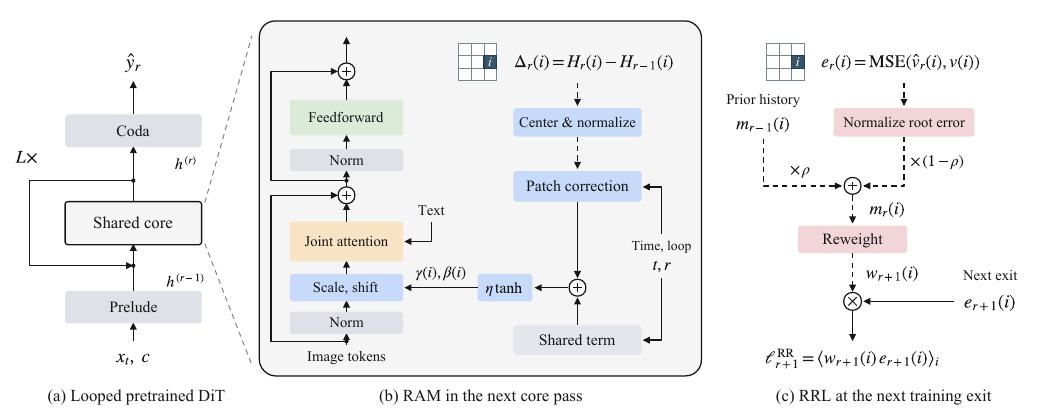}
  \caption{\textbf{LoopFit.}
  (a) The pretrained core repeats at fixed $t$. Training decodes every exit $\hat y_r$. Sampling decodes only the selected exit.
  The marked cells in (b) and (c) trace one position $i$ in computations applied to every image patch.
  (b) From the second pass, Residual Attention Modulation (RAM) adds a correction from the preceding signed feature update at $i$ to the time-and-loop term shared across patches. Their sum sets the scale and shift before the image QKV projections.
  (c) Residual Relay Loss (RRL) compares the velocity derived from $\hat y_r$ with the training target. The history through exit $r$ sets the weight on the new error $e_{r+1}(i)$ at that position. Weights average to one across patches in each image. Dashed signal arrows mark detachment.}
  \label{fig:method}
\end{figure*}

\paragraph{Diffusion models.}
Diffusion, score-based and flow models learn to reverse a noise-corruption process~\cite{ho2020ddpm,song2021sde,lipman2023flow,liu2023rectified}, latent diffusion denoises in a learned autoencoder's latent space~\cite{rombach2022ldm}, and DiT, SiT and JiT denoise with Transformers over latent or pixel patches~\cite{peebles2023dit,ma2024sit,li2025jit}. Unified vision-language models use both discrete and continuous visual representations for image generation~\cite{lu2026unifying}. We retrofit MiniT2I~\cite{minit2i2026}, a pixel-space MMDiT that processes continuous image patches and text tokens with modality-specific weights and joint attention~\cite{esser2024sd3}. These architectures execute a fixed sequence of blocks at each sampling step; T-Stitch, AdaDiff and LoopMDM save sampling compute by running a smaller or shallower network on some steps~\cite{pan2025tstitch,tang2023adadiff,lee2026loopmdm}. For text rendering, DataEvolver uses OCR errors in rejected examples to guide the next round of training data construction~\cite{yan2026dataevolver}. For rule-governed images, SRM denoises the cells of a visual Sudoku in an order the model predicts~\cite{wewer2025srm}, and PaTh steers a frozen diffusion model with a small recursive reasoner, reporting that a recursive denoiser helps its puzzles only marginally~\cite{skiers2026path}. DiffThinker fine-tunes an image editor per task~\cite{he2025diffthinker}. EndoCoT iterates latent states in the text encoder~\cite{dai2026endocot}, while SelfCondFM conditions each flow prediction on the endpoint estimate from the preceding sampling step~\cite{zhang2026selfcondfm}.

\paragraph{Looped Transformers.}
Shared blocks applied repeatedly decouple executed depth from parameter count, for in-context learning~\cite{yang2024looped} and iterative reasoning~\cite{saunshi2025latent}, and at the scale of language models~\cite{geiping2025recurrent}. Pretrained language models have been made recursive with depth-specific low-rank residuals~\cite{bae2025rrt} or a curriculum over recurrences~\cite{mcleish2025retrofitted}, and a vision-language model by looping pruned layers with adapters~\cite{iyengar2026recurrentvlm}; \method reuses the pretrained blocks and adds small controllers for residual attention modulation. Other looped language models change what the shared block reads between loops: the loop input mixed back in~\cite{marchenko2026closing}, parallel residual streams~\cite{zeitoun2026hyperloop} or the decoded state re-embedded as a token~\cite{fu2026discoloop}. Others run the later loops only on the tokens that need them and skip the rest, by halting~\cite{dehghani2019universal,zhu2025ouro,mishra2026dynamic}, routing~\cite{bae2025mor} or a decider, with later loops trained only on a fixed reference model's errors~\cite{fu2026tah}. In our model every patch participates in attention, and RRL changes its weight in the training objective. This relates to residual-based attention~\cite{anagnostopoulos2024rba} and boosting~\cite{yu2023boostnet,song2022boosting}, whereas hard-example mining and the focal loss weight by the current prediction~\cite{shrivastava2016ohem,lin2017focal}.

\paragraph{Residual feedback.}
LRFeed summarizes consecutive latent differences as an unsigned channel-energy profile, pooled across positions, and uses it to modulate normalized attention and FFN features in a looped reasoner~\cite{anonymous2026lrfeed}.
We build on residual-conditioned modulation for pretrained DiT adaptation, retaining signed spatial differences and adding centered patch-dependent corrections to time-and-loop conditioning before image attention.
Residual Relay Loss separately uses target-dependent error history to weight supervision; latent changes alone do not indicate whether an update is correct.

\paragraph{Looped diffusion models.}
Looped-DiT loops blocks inside each diffusion step, supervises every loop through shared output blocks and regulates repeated attention updates with Exclusive Self Attention~\cite{chng2026loopeddit}. LiFT, ELT and LoopREG train looped generators with path targets, intra-loop self-distillation and representation supervision~\cite{derakhshani2026lift,goyal2026elt,wang2026loopreg}, and Correction Training applies a learned corrector repeatedly~\cite{ahn2026correction}. Thinking Pixel adds sparse recursive modules, also to pretrained SD3~\cite{sun2026thinkingpixel}, and Hidden-State Refinement repeats layers of frozen generators without training~\cite{yan2026hiddenstate}. Our retrofit conditions each pass on the preceding change in image features and uses patch errors across exits to supervise the same shared core.

%% file: sec/3_method.tex
\section{Method}
\label{sec:method}

\method reuses a pretrained DiT core several times before taking a denoising step (Fig.~\ref{fig:method}). RAM conditions the next pass on how each patch's hidden state changed. During training, the target supplies a second signal. RRL weights the next exit's loss using the history of prediction errors at the same patch.

\subsection{Retrofitting recurrent computation}
\label{sec:retrofit}

Let $y$ be a clean image and $x_t=t y+(1-t)\sigma\epsilon$ its noisy input, where $t\in[0,1]$ is the clean fraction, $\sigma$ is the noise scale and $\epsilon\sim\mathcal N(0,I)$.
The DiT predicts $y$ given $x_t$ and condition $c$.
We retain its initial blocks as a prelude $P$ and its final blocks and prediction head as a coda $D$.
The intervening blocks form a shared core $C_\theta$. For a given $x_t$, $c$ and $t$, the core updates its state over $L$ passes,
\begin{equation}
\begin{split}
 h^{(0)} &= P(x_t,c),\\
 h^{(r)} &= C_\theta(h^{(r-1)};a_r),\quad
 \hat y_r=D(h^{(r)}).
\end{split}
\label{eq:loop}
\end{equation}
Here $a_r$ collects the scale offsets and shifts that RAM applies on pass $r$, with $a_1=0$.
All three parts inherit their pretrained weights, so the first pass at initialization follows the original computation.
The next pass takes the image and text states in $h^{(r)}$ as input. During training, the shared coda also reads each state to produce the exit prediction $\hat y_r$.

All exits predict the same target at the same noise level.
Following the base velocity objective, let $s_t=\max(1-t,s_{\min})$, $v=(y-x_t)/s_t$ and $\hat v_r=(\hat y_r-x_t)/s_t$.
Let $e_r(i)$ denote the mean squared difference between $\hat v_r$ and $v$ over the pixels and channels in patch $i$.
Our direct conversion, NaiveLoop, trains the unmodulated backbone with uniform deep supervision (DS)~\cite{chng2026loopeddit}.
Its objective is $\sum_r\alpha_r\ell_r$, where $\alpha_r$ is an exit coefficient and $\ell_r=N^{-1}\sum_{i=1}^N e_r(i)$ averages over the image's $N$ patches.
We omit the batch average throughout.

\subsection{Residual Attention Modulation (RAM)}
\label{sec:feature-feedback}

RAM uses the difference between a patch's features before and after a pass to modulate its attention input on the next pass. Following residual-conditioned modulation~\cite{anonymous2026lrfeed}, we apply a learned scale and shift. Diffusion time and pass index supply a term shared across the image, while the signed feature difference supplies a correction specific to each patch.

Let $H_r(i)\in\mathbb R^d$ be the image token at patch $i$ in state $h^{(r)}$, and let $\langle\cdot\rangle_i$ denote a mean over patches of one image.
For each completed pass $r<L$, we center its update channel by channel and normalize by one RMS value per image,
\begin{equation}
\begin{split}
 \Delta_r(i)&=\sg[H_r(i)-H_{r-1}(i)],\\
 d_r(i)&=\Delta_r(i)-\langle\Delta_r\rangle_i,\\
 u_r(i)&=d_r(i)\big/\sqrt{\langle\|d_r\|_2^2\rangle_i/d+\varepsilon}.
\end{split}
\label{eq:feature-residual}
\end{equation}
The stop-gradient operator $\sg$ detaches the feedback branch. Gradients can still flow through the recurrent state in Eq.~\ref{eq:loop}.

A shared patch encoder combines $u_r(i)$ with Fourier embeddings $\phi$ of time and completed pass index,
\begin{equation}
\begin{split}
 q_r(i)&=\operatorname{SiLU}\!\left(W[u_r(i);\phi(t);\phi(r/(L-1))]+b\right),\\
 \widetilde q_r(i)&=q_r(i)-\langle q_r\rangle_i.
\end{split}
\label{eq:feature-encode}
\end{equation}
Centering makes the patch correction average to zero before the nonlinearity. Each core block $j$ adds this correction to its own time-and-pass term $b_j(t,r)$ to set the modulation for pass $r+1$,
\begin{equation}
 [\gamma_{r+1,j}(i);\beta_{r+1,j}(i)]
 =\eta\tanh\!\left(b_j(t,r)+A_j\widetilde q_r(i)\right).
\label{eq:feature-gates}
\end{equation}
The scale and shift act on the normalized image token $z$ immediately before its query, key and value projections,
\begin{equation}
 z'=(1+\gamma_{r+1,j})\odot z+\beta_{r+1,j}.
\label{eq:feature-injection}
\end{equation}
The text stream keeps its original normalization and participates in joint attention.
The bound $\eta$ limits the modulation amplitude. Zero-initializing each bias-free projection $A_j$ makes the initial patch correction zero.

\input{tables/reasoning-generation}

\subsection{Residual Relay Loss (RRL)}
\label{sec:focus}

During training, each exit predicts the same clean image, so we can track the error at the same patch across passes. Inspired by cumulative residual weighting~\cite{anagnostopoulos2024rba}, RRL uses this history to weight the patch's loss at the next exit.
After exit $r$, we divide its detached root patch errors $\sg[\sqrt{e_r(i)}]$ by their maximum within the image to obtain $p_r(i)$, taking $p_r=0$ when all errors vanish.
Starting with $m_0=0$, we update
\begin{equation}
 m_r=\rho m_{r-1}+(1-\rho)p_r,\qquad
 \bar m_r=\frac{m_r}{1-\rho^r}.
\label{eq:relay-memory}
\end{equation}
Here $\bar m_r$ is the bias-corrected average. At $\rho=0$ it contains only the latest error profile. For $0<\rho<1$, earlier profiles contribute with geometrically decreasing weight.
Exit $r$ uses only the history through exit $r-1$. With $w_1(i)=1$, we set
\begin{equation}
\begin{split}
 w_r(i)&=\frac{(1+\bar m_{r-1}(i))^2}
 {\langle(1+\bar m_{r-1})^2\rangle_i}\quad(r\geq 2),\\
 \ell_r^{\mathrm{RR}}&=\langle w_r e_r\rangle_i,\qquad
 \mathcal L_{\mathrm{RR}}=\sum_{r=1}^{L}\alpha_r\ell_r^{\mathrm{RR}}.
\end{split}
\label{eq:focus}
\end{equation}
The offset keeps every patch supervised. Mean-one normalization redistributes weight within each image and exit. The error history is discarded after each training example and is never needed for sampling.

\paragraph{Adaptation.}
We first train the recurrent backbone with uniform patch weights for 500 updates, then linearly increase the contribution of relay weighting to full strength at update 1,500.
With the backbone frozen, we fit the shared term of RAM, then hold it fixed while learning the patch correction with Eq.~\ref{eq:focus}.

\subsection{Noise-dependent depth at inference}
\label{sec:noise-depth}

The sampler can spend more recurrent passes early in denoising, when the input is mostly noise. Our fixed schedule uses four passes for the first 40\% of steps and one thereafter. Over 20 Euler steps this uses 44 core passes, compared with 80 at fixed depth four.

\paragraph{Residual-based early exit.}
An image may also exit before the scheduled cap. After pass $r$, we measure the magnitude of each raw patch update, before RAM's centering and normalization. We divide its channelwise RMS by the RMS of the incoming token $H_{r-1}(i)$ plus a small stabilizer, then average the largest tenth of these ratios. This retains sensitivity to changes confined to a few patches. The image exits when the score falls at or below $\tau$ or it reaches the cap. We select $\tau$ on 96 held-out boards per task to minimize the mean pass count while solving at most one fewer board than the fixed schedule.

Each Euler step starts with fresh features and an unmodulated pass. Under early exit, RAM encodes the pass index with $L=4$ in Eq.~\ref{eq:feature-encode}, even for images that stop sooner. The sampler decodes each image once at its chosen exit and takes one Euler step.

%% file: tables/reasoning-generation.tex
\begin{table*}[t]
  \centering
  \small
  \renewcommand{\arraystretch}{1.00}
  \setlength{\tabcolsep}{4.2pt}
  \begin{tabular}{l r *{13}{c}}
    \toprule
    & & \multicolumn{3}{c}{FrozenLake} & \multicolumn{3}{c}{Maze} & \multicolumn{3}{c}{TSP} & \multicolumn{3}{c}{Sudoku} & \\
    \cmidrule(lr){3-5}\cmidrule(lr){6-8}\cmidrule(lr){9-11}\cmidrule(lr){12-14}
    Model & Params & 16 & 32 & Hard & 16 & 32 & Hard & 12 & 15 & Hard & 40 & 35 & Hard$^{\dagger}$ & Avg. \\
    \midrule
    \multicolumn{15}{l}{\textit{Zero-shot}} \\
    GPT-5 & -- & 3 & 0 & 0 & 0 & 0 & 0 & 0 & 0 & 0 & 0 & 0 & 1 & 0.3 \\
    Gemini-3-Flash & -- & 52 & 3 & 0 & 0 & 0 & 0 & 25 & 9 & 4 & 29 & 3 & 0 & 10.4 \\
    Qwen3-VL-32B~\cite{bai2025qwen3vl} & 32B & 0 & 0 & 0 & 0 & 0 & 0 & 0 & 0 & 0 & 0 & 0 & 0 & 0.0 \\
    Qwen-Image-Edit-2511~\cite{wu2025qwenimage} & 20B & 0 & 0 & 0 & 0 & 0 & 0 & 0 & 0 & 0 & 0 & 0 & 0 & 0.0 \\
    ThinkGen~\cite{jiao2025thinkgen} & 4B & 0 & 0 & 0 & 0 & 0 & 0 & 0 & 0 & 0 & 0 & 0 & 0 & 0.0 \\
    ChronoEdit~\cite{wu2025chronoedit} & 14B & 1 & 1 & 0 & 1 & 26 & 0 & 0 & 0 & 0 & 0 & 0 & 0 & 2.4 \\
    \midrule
    \multicolumn{15}{l}{\textit{Task-specific training}} \\
    Qwen3-VL-32B (SFT) & 32B & 85 & 21 & -- & 57 & 3 & -- & 69 & 59 & 52 & 22 & 2 & -- & -- \\
    Qwen3-VL-32B (GRPO) & 32B & 1 & 0 & -- & 0 & 0 & -- & 0 & 0 & 0 & 1 & 0 & -- & -- \\
    DiffThinker~\cite{he2025diffthinker} & 20B & 96 & 83 & 56 & 97 & 56 & 1 & 74 & 62 & 58 & 95 & 57 & 22 & 63.1 \\
    DiffThinker++~\cite{he2025diffthinker} & 20B & 99 & 80 & 48 & \textbf{100} & 65 & 7 & 76 & 72 & 59 & 94 & 55 & 21 & 64.7 \\
    EndoCoT~\cite{dai2026endocot} & 20B & 99 & 85 & -- & \textbf{100} & 90 & 80 & 77 & 77 & \textbf{73} & \textbf{100} & 95 & 20 & -- \\
    SelfCondFM~\cite{zhang2026selfcondfm} & 20B & \textbf{100} & 86 & 83 & \textbf{100} & 95 & 85 & \textbf{79} & 74 & 72 & \textbf{100} & 98 & \textbf{82} & 87.8 \\
    \midrule
    \rowcolor[rgb]{0.90,0.94,0.96}
    \textbf{\method (ours)} & 0.26B & \textbf{100} & \textbf{87} & \textbf{91} & \textbf{100} & \textbf{100} & \textbf{94} & 75 & \textbf{78} & \textbf{73} & \textbf{100} & \textbf{99} & 80 & \textbf{89.8} \\
    \bottomrule
  \end{tabular}
  \caption{\textbf{Reasoning by generation.} Success (\%) on 100 puzzles per setting. Hard uses FrozenLake boards with 100-move shortest paths, independently generated long-path Maze-32 boards and released 18-city TSP instances. $^{\dagger}$Sudoku Hard uses the released 30-given puzzles re-rendered with our training typeface and layout. Avg.\ covers all 12 settings. Dashes denote unavailable scores. Each \method model uses 20 Euler steps and four passes per step.}
  \label{tab:puzzleref}
\end{table*}

%% file: sec/4_experiments.tex
\section{Experiments}
\label{sec:exp}

\subsection{Main Results}
\label{sec:puzzles}

We evaluate \method on four tasks that require drawing the solution to a visual puzzle~\cite{he2025diffthinker}. Across the 12 settings in Table~\ref{tab:puzzleref}, \method solves 89.8\% of puzzles, compared with 87.8\% for SelfCondFM, the strongest baseline with results on all settings.

\paragraph{Longer paths at the same grid size.}
On released Maze-32, \method solves all 100 boards, SelfCondFM solves 95 and EndoCoT solves 90. On Maze Hard, the gap widens. \method solves 94 of the 100 long-path boards, compared with 85 for SelfCondFM and 80 for EndoCoT. DiffThinker++ solves seven and DiffThinker solves one.
The Maze Hard paths have a median length of 609 moves, compared with 202 on released Maze-32, at the same grid size. Success requires an unbroken route from start to goal, so a local break can invalidate a largely correct drawing.
On FrozenLake, \method reaches the attainable maxima of 100 and 87 on the two released settings. On FrozenLake Hard, it solves 91 boards, eight more than SelfCondFM and 43 more than DiffThinker++. Together, Maze Hard and FrozenLake Hard account for 17 of the 23 additional solutions over SelfCondFM across all 12 settings.

\paragraph{Optimal tours and completed grids.}
On TSP, \method finds 75, 78 and 73 optimal tours at 12, 15 and 18 cities. It leads by one tour at 15 cities. At 18 cities, it ties EndoCoT and solves one more board than SelfCondFM. On Sudoku, \method solves all 100 puzzles with 40 givens and 99 with 35 givens. On the re-rendered 30-given puzzles it solves 80, two fewer than SelfCondFM. DiffThinker++ and EndoCoT solve 21 and 20, respectively.

\begin{figure*}[t]
  \centering
  \includegraphics[width=\textwidth]{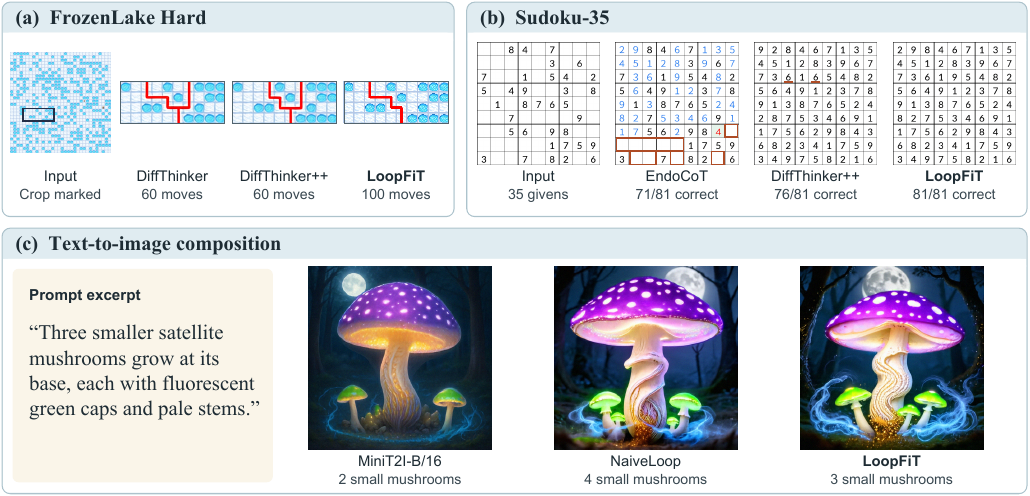}
  \caption{\textbf{Puzzle solving and text-to-image composition.} (a) The crops show the boxed region. Both DiffThinker variants stop after 60 moves on the full board, while \method reaches the goal in 100. (b) Boxes mark empty cells and underlines mark repeated sixes. (c) The prompt specifies three small green-capped mushrooms at the base of a large purple one. Only \method matches that count in this example.}
  \label{fig:puzzle-cases}
\end{figure*}

In Figure~\ref{fig:puzzle-cases}a, \method draws a continuous route through a region where both DiffThinker variants draw a fork. On Sudoku-35 (Figure~\ref{fig:puzzle-cases}b), \method recovers the full solution. DiffThinker++ fills all 81 cells but repeats a six in the third row, and EndoCoT leaves ten cells empty.

\paragraph{Does fine-tuning at the original depth suffice?}
On a separate set of 384 long-path mazes, we compare two backbones initialized from the same MiniT2I model and trained on the same 30,000 mazes. At matched estimated training FLOPs, the single-pass backbone solves 164 boards after 23,167 updates and the four-pass backbone trained with RRL solves 293 after 8,000. Both use 20 Euler steps and neither uses an adapter. At the training budget of full \method, 38 single-pass steps use nearly as many block calls as 20 four-pass steps (646 versus 640). The single-pass model then solves 165 boards, compared with 366 for \method.

\paragraph{Transfer to text-to-image generation.}
We also adapt MiniT2I-B/16 to generate images from text prompts. T2I-CoReBench~\cite{li2025corebench} tests whether an image satisfies explicit requirements (Figure~\ref{fig:puzzle-cases}c) and whether it depicts content that must be inferred from the prompt. \method raises the overall score from 57.09 for NaiveLoop to 58.55 (Table~\ref{tab:t2i-transfer}). Both composition and reasoning improve over NaiveLoop. On reasoning, \method scores 49.40, slightly below the pretrained model's 49.77. On English text rendering in OneIG~\cite{chang2025oneig}, \method scores 15.53 versus 14.02 for NaiveLoop. The unweighted mean over CoRe, PRISM Overall~\cite{fang2025prism}, SpatialGenEval~\cite{wang2026spatialgeneval}, DPG-Bench~\cite{hu2024ella} and OneIG rises from 53.53 to 54.13.

\begin{table}[ht]
  \centering
  \small
  \setlength{\tabcolsep}{5.8pt}
  \begin{tabular}{@{}lrrr@{}}
    \toprule
    Model & Composition & Reasoning & Overall \\
    \midrule
    MiniT2I-B/16 & 68.46 & \textbf{49.77} & 56.00 \\
    NaiveLoop & 75.68 & 47.80 & 57.09 \\
    \rowcolor[rgb]{0.90,0.94,0.96}
    \method{} & \textbf{76.86} & 49.40 & \textbf{58.55} \\
    \bottomrule
  \end{tabular}
  \caption{\textbf{Composition and reasoning on T2I-CoReBench.} The overall score averages all 12 tasks, four for composition and eight for reasoning. All three denoisers have 0.26B parameters.}
  \label{tab:t2i-transfer}
\end{table}

\subsection{Ablations of RAM and RRL}
\label{sec:feedback-eval}

At four passes without adapters, the 3K RRL backbone solves 49 of 384 long-path mazes versus 25 for the uniformly supervised backbone, gaining 30 boards and losing six. The runs share the initialization, training set and update budget, but differ in GPU count, kernel compilation and training draws.

For the adapter ablations in Table~\ref{tab:feedback-current}, all variants within a task share a frozen RRL-trained backbone and the same initial time-and-loop controller. We match training examples and noise draws within each seed. Regular and hard Sudoku use the same typeface, with fewer givens on the hard boards.

\input{tables/feedback-adaptation}

\paragraph{Spatial alignment of the feature residual.}
With RRL fixed, adding the patch residual raises success from 67 to 110 on the 192 hard Maze development boards. It solves 45 boards missed by the time-and-loop control and loses two of its successes. The gain is one board on regular Maze and averages 3.08 percentage points on hard Sudoku.
See2Think probes whether multimodal models use intermediate images by corrupting the visual evidence they receive~\cite{yan2026see2think}. We test whether RAM uses each residual's location by shifting the residuals by half the patch sequence at inference in the same trained Maze model. The model then solves 30 of the 192 boards, compared with 53 when the patch branch is disabled and 110 when each patch receives its own residual.

\paragraph{How much error history is useful?}
On hard Maze, RRL gains one board over uniform weighting with RAM off and 30 with RAM on (Table~\ref{tab:feedback-current}a).

To isolate the contribution of older errors, we keep the patch branch enabled and vary the history used to weight its loss (Table~\ref{tab:feedback-current}b).
On 384 independent long-path mazes, uniform weights, the preceding pass alone, and cumulative history yield 162, 180, and 202 solutions. Cumulative weighting preserves all 180 solutions found with the preceding pass and solves 22 more.
On hard Sudoku, extending the history raises mean success from 53.08\% to 53.50\%. It adds five net solutions for one seed and leaves the total unchanged for the other. On regular Sudoku, uniform weights with the patch branch give the highest mean score (99.00\%).

\begin{figure*}[t]
  \centering
  \includegraphics[width=\textwidth]{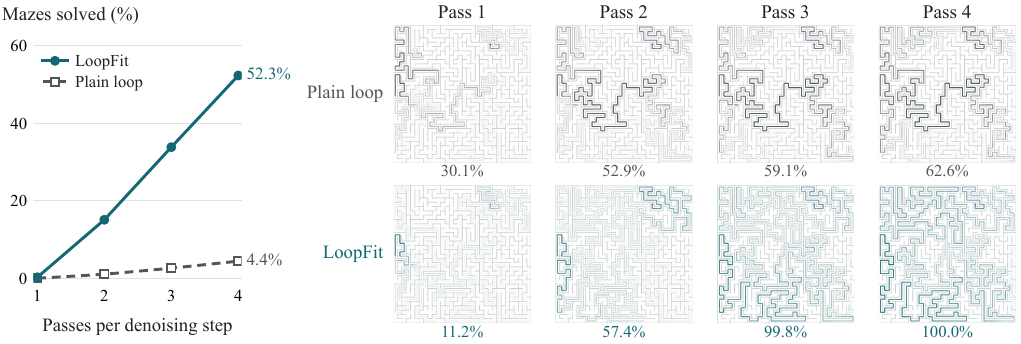}
  \caption{\textbf{Recurrent refinement on Hard Maze.} The curves show final success on 384 mazes with depth fixed across 20 Euler steps. The images show the clean prediction after each pass of a single four-pass forward through each model at the first denoising step. Both models receive the same maze and initial noise. Percentages below the images give path IoU, scored before recoloring the predictions to match the curves.}
  \label{fig:maze-depth}
\end{figure*}

\paragraph{Results on released boards.}
On the released FrozenLake and Sudoku settings, the shared controller, patch modulation with uniform supervision, and patch modulation with RRL give the same success counts after 400 adaptation updates. On released TSP, the 28K backbone finds 222 optimal tours. With 400 matched updates, the time-and-loop controller finds 219 and the patch residual variant finds 223, gaining eight tours and losing four relative to the controller.

%% file: tables/feedback-adaptation.tex
\begin{table}[t]
\centering
\small
\setlength{\tabcolsep}{5pt}
\renewcommand{\arraystretch}{1.13}
\begin{tabular}{*{2}{w{c}{\dimexpr\columnwidth/6-2\tabcolsep\relax}}*{4}{w{r}{\dimexpr\columnwidth/6-2\tabcolsep\relax}}}
\toprule
\multicolumn{6}{@{}l}{\textbf{(a) Component ablation} \quad Success rate (\%)} \\
\addlinespace[2pt]
\multicolumn{2}{c}{\textbf{Adaptation}} & \multicolumn{2}{c}{\textbf{Maze}} & \multicolumn{2}{c}{\textbf{Sudoku}} \\
\cmidrule(lr){1-2}\cmidrule(lr){3-4}\cmidrule(lr){5-6}
RAM & RRL & Regular & Hard & Regular & Hard \\
\midrule
-- & -- & 73.44 & 34.38 & 98.33 & 48.33 \\
$\checkmark$ & -- & 71.88 & 41.67 & \textbf{99.00} & 52.00 \\
-- & $\checkmark$ & 73.44 & 34.90 & 98.33 & 50.42 \\
\rowcolor[rgb]{0.90,0.94,0.96}
$\checkmark$ & $\checkmark$ & \textbf{75.00} & \textbf{57.29} & 98.67 & \textbf{53.50} \\
\bottomrule
\end{tabular}\par
\vspace{4pt}
\begin{tabular}{w{l}{\dimexpr0.48\columnwidth-2\tabcolsep\relax}*{2}{w{r}{\dimexpr0.26\columnwidth-2\tabcolsep\relax}}}
\toprule
\multicolumn{3}{@{}l}{\textbf{(b) Error history} \quad Success rate (\%)} \\
\addlinespace[2pt]
Training loss & Hard Maze & Hard Sudoku \\
\midrule
Uniform patch weights & 42.19 & 52.00 \\
Previous pass ($\rho=0$) & 46.88 & 53.08 \\
\rowcolor[rgb]{0.90,0.94,0.96}
Cumulative RRL ($\rho=0.9$) & \textbf{52.60} & \textbf{53.50} \\
\bottomrule
\end{tabular}
\caption{\textbf{Residual feedback during adaptation.} (a) With RAM off we fit shared modulation. With RAM on we freeze it and fit the patch correction. RRL off uses uniform patch weights. Backbone, shared-controller and adaptation budgets are 3K/120/400 updates for Maze (64 regular, 192 hard boards, one seed) and 10K/120/3K for Sudoku (300 regular, 600 hard, two-seed mean). (b) With RAM on, we vary the error history on 384 independent long-path mazes and the same hard Sudoku boards.}
\label{tab:feedback-current}
\end{table}

%% file: sec/5_analysis.tex
\subsection{Analysis of Recurrent Refinement}
\label{sec:feedback-analysis}

\begin{figure}[tb]
  \centering
  \includegraphics[width=\columnwidth]{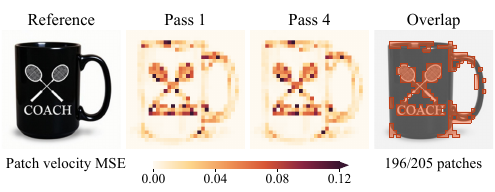}
  \caption{\textbf{High-error patches persist across passes.} The mug comes from the FLUX-Reason text-rendering subset~\cite{fang2025prism}. The error maps share a color scale. Orange outlines patches in the highest-error fifth at both passes, tracing the lettering, crossed rackets and parts of the handle.}
  \label{fig:t2i-patch-persistence}
\end{figure}

\paragraph{Where does prediction error persist?}
\label{sec:where}
We evaluate the uniformly supervised 20K NaiveLoop on 48 general images from the training mix, holding each noisy input fixed at $t=0.45$ across passes. We rank the 1,024 patches in each image by first-pass velocity MSE. Of the top 205, an average of 93.6\% remain in the highest-error fifth at pass four. Together, the patches selected at pass one account for an average of 62.6\% of the fourth-pass error. RRL uses this persistence to give more weight in later passes to patches with a history of large errors. Figure~\ref{fig:t2i-patch-persistence} illustrates this overlap on a separate text-rendering example.

\paragraph{What does another recurrent pass buy?}
We compare \method with a plain loop trained with uniform patch weights throughout. Both recipes see the same number of training examples. For the final 520 updates, the plain loop continues to train its backbone while \method fits its adapters. At depths one through four, \method solves 1, 58, 130 and 201 mazes, versus 0, 4, 10 and 17 for the plain loop (Fig.~\ref{fig:maze-depth}). At four passes, \method solves 186 mazes missed by the plain loop and loses two of its successes. On the illustrated maze, its second pass marks 382 of the 665 route cells, but the path from the start breaks after 18 moves. The third pass fills the gaps and connects the entire 664-move route.

\paragraph{Which patches receive more weight?}
On 192 hard Maze boards, we ask how much weight earlier passes assign to the fifth of patches with the largest fourth-pass errors. We evaluate both weighting rules on the same frozen 3K backbone at $t=0.5$. These patches receive 20.02\% of the weight under uniform supervision, 22.82\% using the previous pass, and 22.65\% using cumulative history. The two histories thus allocate a similar share to this group, even though the adapter trained with cumulative history solves more long-path mazes (Table~\ref{tab:feedback-current}b).

\begin{table}[tb]
  \centering
  \small
  \setlength{\tabcolsep}{0pt}
  \begin{tabular}{@{}w{l}{.34\columnwidth}*{4}{w{c}{.165\columnwidth}}@{}}
    \toprule
    & \multicolumn{2}{c}{4 training passes} & \multicolumn{2}{c}{8 training passes} \\
    \cmidrule(lr){2-3}\cmidrule(lr){4-5}
    Inference passes & 4 & 8 & 4 & 8 \\
    \midrule
    Maze (384)   & 50 & 45 & 35 & \textbf{82} \\
    Sudoku (600) & 48 & \textbf{86} & 16 & 70 \\
    \bottomrule
  \end{tabular}
  \caption{\textbf{Training and inference depth.} Entries count solved boards on long-path Maze-32 and hard Sudoku. All RRL backbones train for 3K updates. Eight passes at training time cost more per update than four.}
  \label{tab:maze-train-infer-depth}
\end{table}

\paragraph{Does training at greater depth help?}
With four training passes (Table~\ref{tab:maze-train-infer-depth}), doubling inference depth solves five fewer mazes (6 gained, 11 lost) and 38 more Sudoku puzzles (40 gained, 2 lost). At a fixed inference depth of eight, increasing training depth from four to eight has the opposite effect. It solves 37 more mazes (51 gained, 14 lost) and 16 fewer Sudoku puzzles (36 gained, 52 lost).

\paragraph{When does patchwise feedback help?}
RAM and the time-and-loop control share a frozen 8K Maze backbone, RRL objective, adaptation budget and initial noise (Table~\ref{tab:maze-depth-pairs}). Both use 20 Euler steps at fixed depth. At one pass, the feedback branch is inactive and all 384 output pairs are pixel-identical. At pass two, each patch first receives its preceding feature update alongside the shared time-and-loop signal. RAM gains 84 solutions and loses nine. The control solves 78 of those 84 boards at four passes, showing that RAM recovers many solutions at shallower depth. At four passes, RAM retains all 347 control successes and solves 19 more, a gain of 4.95 percentage points.

\begin{table}[t]
  \centering
  \small
  \setlength{\tabcolsep}{2pt}
  \begin{tabular*}{\columnwidth}{@{\extracolsep{\fill}}r rr rr rr@{}}
    \toprule
    & \multicolumn{2}{c}{Time + loop} & \multicolumn{2}{c}{With RAM} & \multicolumn{2}{c}{RAM vs.\ control}\\
    \cmidrule(lr){2-3}\cmidrule(lr){4-5}\cmidrule(l){6-7}
    $L$ & Solved & $+/-$ & Solved & $+/-$ & Gained & Lost\\
    \midrule
    1 &  23 & ---   &  23 & ---   &  0 & 0\\
    2 & 201 & 179/1 & 276 & 255/2 & 84 & 9\\
    3 & 292 & 97/6  & 337 & 66/5  & 50 & 5\\
    4 & 347 & 58/3  & 366 & 30/1  & 19 & 0\\
    \bottomrule
  \end{tabular*}
  \caption{\textbf{Patch feedback at matched inference depth.} Both adapters use the same 8K backbone and the same 384 long-path mazes. Within each model, $+/-$ counts solutions gained or lost from depth $L-1$ to $L$. The last two columns compare RAM with the time-and-loop control at fixed $L$.}
  \label{tab:maze-depth-pairs}
\end{table}

\begin{figure}[tb]
  \centering
  \includegraphics[width=\columnwidth]{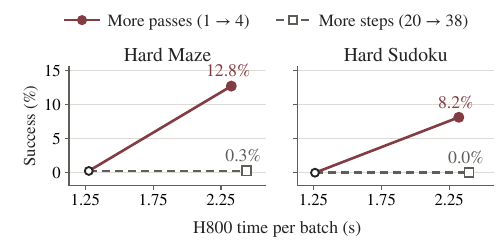}
  \caption{\textbf{Where should extra inference compute go?} Both comparisons start from one pass per step and 20 Euler steps (open circle). Each point uses the same boards and initial noise within a task. Time is the median per batch of 16 after warm-up.}
  \label{fig:loop-vs-denoise}
\end{figure}

\paragraph{More denoising steps or deeper predictions?}
We vary inference on the same four-pass backbone per task (3K updates with RRL, before RAM adaptation). From a budget of 340 block calls, we can take 38 single-pass Euler steps (646 calls) or keep 20 steps and run four passes per step (640 calls). The two choices take 2.3--2.4\,s per batch (Fig.~\ref{fig:loop-vs-denoise}). Extra steps leave every board's outcome unchanged. Extra passes solve 48 additional mazes out of 384 and 49 additional Sudoku puzzles out of 600, retaining every baseline success.

\begin{figure}[t]
  \centering
  \includegraphics[width=\columnwidth]{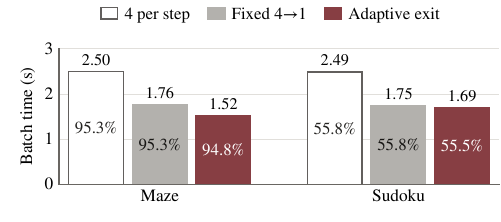}
  \caption{\textbf{Allocating recurrent passes at inference.} Bars show median H800 time per batch of 16. Labels inside the bars give success rates on 384 mazes and 600 Sudoku puzzles (seed 43 for Sudoku). All policies use 20 Euler steps and the same initial noise. Exit thresholds are 0.331 for Maze and 0.105 for Sudoku.}
  \label{fig:depth-allocation}
\end{figure}

\paragraph{When do extra passes help most?}
With 44 passes over 20 Euler steps, moving the eight four-pass steps from the end to the start raises success from 23/384 to 366/384 on Maze (H800) and from 18/600 to 335/600 on Sudoku (B300). Placing the extra passes at high noise gains 343 mazes and 317 Sudoku puzzles and loses none. On Maze, spreading the same 44 passes nearly evenly across steps solves 279/384. Early allocation gains 89 boards and loses two against this policy. These controls favor reserving recurrent refinement for the high-noise portion of denoising.

Fixed $4{\to}1$ matches four passes throughout (95.3\% on Maze, 55.8\% on Sudoku) and cuts median batch time by 29.8\% on both (Fig.~\ref{fig:depth-allocation}). Residual-based early exit adapts this budget to each image, using feature-update magnitude within the scheduled cap. It saves another 13.7\% on Maze and 3.3\% on Sudoku at costs of 0.52 and 0.33 percentage points in success rate. Against fixed schedules at matched pass budgets, it gains two boards per task, taking 5.5\% longer on Maze and 1.5\% longer on Sudoku.

%% file: sec/6_conclusion.tex
\section{Conclusion}
\label{sec:conclusion}

\method equips a pretrained DiT with patchwise feedback for recurrent refinement. RAM conditions attention on each patch's preceding feature update, while RRL weights supervision by accumulated prediction errors. In visual reasoning, the largest gains over SelfCondFM occur on long paths in Maze and FrozenLake. In text-to-image generation, the recurrent retrofit improves English text rendering over NaiveLoop. RAM recovers solutions in fewer passes and adds solutions at full depth. Cumulative error history improves hard-Maze adaptation, with smaller gains on hard Sudoku. Allocating depth to high-noise steps maintains success rates and reduces sampling time on both tasks.

%% file: sec/7_llm_statement.tex
\section*{LLM Usage Statement}
We used OpenAI ChatGPT and Codex to help design controls, implement and troubleshoot experiments, summarize per-board results, and revise the figures and manuscript.